\documentclass{article}

\usepackage[preprint]{neurips_2023}

\usepackage[utf8]{inputenc} 
\usepackage[T1]{fontenc}    
\usepackage{hyperref}       
\usepackage{url}            
\usepackage{booktabs}       
\usepackage{amsfonts}       
\usepackage{nicefrac}       
\usepackage{microtype}      
\usepackage{xcolor}         
\usepackage{graphicx}

\usepackage{inconsolata}
\usepackage{amsmath}
\usepackage{amssymb}
\usepackage{listings}
\usepackage{xcolor}       
\usepackage{multirow}

\makeatletter
{\small 
\xdef\f@size@small{\f@size}
\xdef\f@baselineskip@small{\f@baselineskip}
\normalsize 
\xdef\f@size@normalsize{\f@size}
\xdef\f@baselineskip@normalsize{\f@baselineskip}
}
\newcommand{\smalltonormalsize}{%
  \fontsize
    {\fpeval{(\f@size@small+\f@size@normalsize)/2}}
    {\fpeval{(\f@baselineskip@small+\f@baselineskip@normalsize)/2}}%
  \selectfont
}
\makeatother

\newcommand{\botname}{Higgs }

\title{Code Soliloquies for Accurate Calculations in \\
Large Language Models}

\author{Shashank Sonkar \\
  Rice University \\
  \texttt{ss164@rice.edu} \\\And
  MyCo Le \\
  Rice University \\
  \texttt{msl12@rice.edu} \\\And
  Xinghe Chen \\
  Rice University \\
  \texttt{xc42@rice.edu} \\\AND
  Naiming Liu \\
  Rice University \\
  \texttt{nl35@rice.edu} \\\And
  Debshila Basu Mallick \\
  OpenStax \\
  \texttt{db19@rice.edu} \\\And
  Richard G. Baraniuk \\
  Rice University \\
  \texttt{richb@rice.edu}
  }
\begin{document}

\maketitle
\begin{abstract}

High-quality conversational datasets are crucial for the successful development of Intelligent Tutoring Systems (ITS) that utilize a Large Language Model (LLM) backend. Synthetic student-teacher dialogues, generated using advanced GPT-4 models, are a common strategy for creating these datasets. However, subjects like physics that entail complex calculations pose a challenge. While GPT-4 presents impressive language processing capabilities, its limitations in fundamental mathematical reasoning curtail its efficacy for such subjects. To tackle this limitation, we introduce in this paper an innovative stateful prompt design. Our design orchestrates a mock conversation where both student and tutorbot roles are simulated by GPT-4. Each student response triggers an internal monologue, or `code soliloquy' in the GPT-tutorbot, which assesses whether its subsequent response would necessitate calculations. If a calculation is deemed necessary, it scripts the relevant Python code and uses the Python output to construct a response to the student. Our approach notably enhances the quality of synthetic conversation datasets, especially for subjects that are calculation-intensive. Our preliminary Subject Matter Expert evaluations reveal that our Higgs model, a fine-tuned LLaMA model, effectively uses Python for computations, which significantly enhances the accuracy and computational reliability of Higgs' responses. Code, models, and datasets is available at \url{https://github.com/luffycodes/Tutorbot-Spock-Phys}.
\end{abstract}

\section{Introduction}
\begin{figure*}[ht!]
    \centering
    \includegraphics[width=\textwidth]{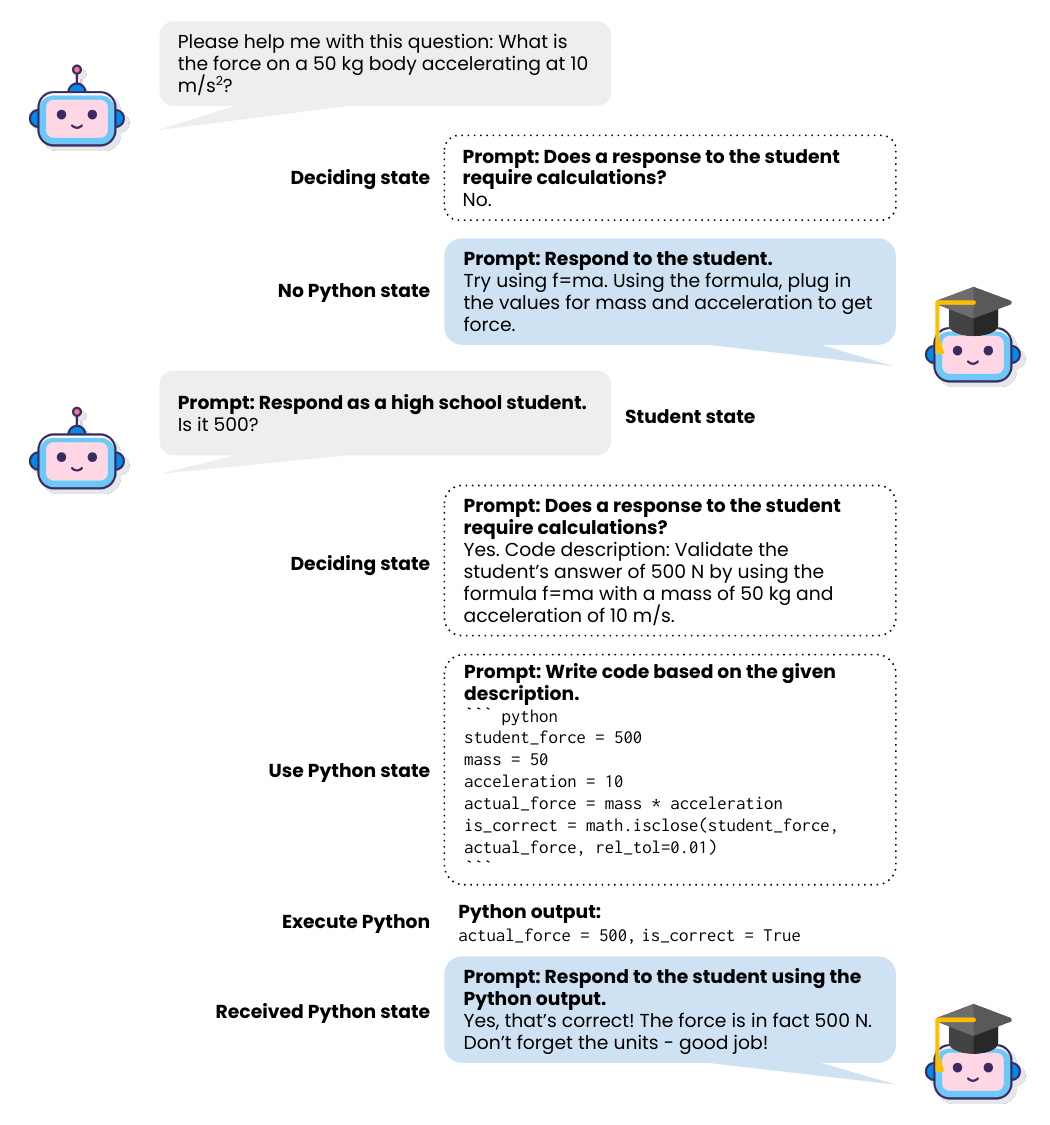}
    \caption{
    An example of a synthetic student-tutorbot conversation generated using our proposed multi-turn, stateful prompt design.
    Both the student and the tutorbot roles are simulated by the GPT model.
    The goal of our prompt design is to ensure the mathematical accuracy of responses from the GPT-tutorbot, such as in scenarios that require calculation verification or responses to calculation-based queries.
    To achieve this, our design engages the tutorbot in an `internal monologue', what we term a `code soliloquy'.
    This soliloquy, illustrated by the dotted bubbles in the figure, is hidden from the student.
    The soliloquy is initiated each time the tutorbot receives a student input and is guided by a sequence of four state prompts.
    The first state in this soliloquy prompts the tutorbot to assess whether the next response necessitates any calculations.
    If the answer is affirmative, the tutorbot is prompted to generate the corresponding Python code, and the output of this code is then utilized to formulate the tutorbot's response. 
    In contrast, if the tutorbot determines that Python is not required, it proceeds to respond without invoking any Python.
    The prompts in the figure are greatly simplified for illustrative purposes.
    Detailed versions of these prompts can be found in the appendix.
    }
    \label{fig:mock_con}
\end{figure*}

In the rapidly evolving domain of Natural Language Processing (NLP), the creation of high-quality chatbots using pre-trained Large Language Models (LLMs) is heavily reliant on conversational datasets as shown by Vicuna model \citep{vicuna2023}.
With advanced models like Generative Pretrained Transformer-4 (GPT-4) \citep{gpt4sparkai}, it is possible to generate such synthetic yet engaging conversations by designing creative prompts \citep{sonkar2023class}.
CLASS framework \citep{sonkar2023class} demonstrates the capacity of GPT-4 to synthesize meaningful interactions between a student and a tutorbot to train effective Intelligent Tutoring Systems (ITS).
However, this framework largely caters to subjects that circumvent calculation-intensive problems, such as biology.

Generating synthetic conversations for subjects like physics, which require complex calculations, presents a significant challenge.
This is primarily due to the limited mathematical capabilities of models like GPT-4.
For instance, ChatGPT \cite{kasneci2023chatgpt} and GPT-4 achieve only $55\%$ and $59\%$ accuracy respectively on three-digit by three-digit multiplication tasks, as reported by \citep{gptbadmath}.
This limitation makes the one-shot prompt design introduced in CLASS inadequate for generating holistic conversations in calculation-intensive subjects. 

Recognizing these limitations in GPT-4's mathematical capabilities, we have developed an innovative approach to generate synthetic student-tutor dialogues (example shown in figure~\ref{fig:mock_con}) that incorporate accurate calculations.
Our solution is a multi-turn, stateful prompt design that leverages GPT-4 to simulate both student and tutorbot roles.
Central to this design is the unique incorporation of `code soliloquies', a novel concept that significantly enhances the dialogue's computational accuracy.
For every input from the GPT-4 emulated student, we initiate a soliloquy within the GPT-4 tutorbot – an internal dialogue hidden from the student. 
During  this soliloquy, GPT-tutorbot prompts itself to determine whether its next response necessitates any calculations.
If a calculation is required, it proceeds to script the necessary code and then utilizes the output from this code to generate an informed response to the student. 
Given that GPT-4 demonstrates a remarkable proficiency in writing code, we ingeniously utilize this strength in our design through the process of code soliloquy. 
This allows us to overcome GPT-4's calculation limitations, thereby significantly enhancing the quality of the synthetic dialogues, particularly for subjects that are calculation-intensive.

To demonstrate the efficacy of our stateful prompt design and the quality of the generated synthetic conversations, we introduce Higgs, a variant of the LLaMA-2-70b-chat base model \citep{llama2}, fine-tuned on the generated conversational dataset.
The starting question posed by the GPT-student to the GPT-tutorbot to initiate these conversations is our newly curated physics question dataset, PHY300, adapted from high school physics textbooks.
These questions are carefully chosen to cover a broad spectrum of topics, ranging from mechanics to thermodynamics to electromagnetism.

In order to test Higgs, we develop a comprehensive evaluation protocol.
The evaluation measures the accuracy and computational reliability of Higgs' responses, particularly its proficiency in using Python for computations whenever necessary.
The results from preliminary SME evaluations are highly encouraging. 
Higgs exhibited an impressive ability to determine when Python computations were necessary in the conversation, and it consistently generated valid code.
Most notably, Higgs accurately verified student's calculations by leveraging Python code, underlining the utility of our approach in improving the computational reliability of LLMs.
These results demonstrate the potential of our stateful prompt design and generated mock conversations with code soliloquies in significantly enhancing the capabilities of LLMs, particularly in the context of calculation-intensive subjects. 
By fostering accuracy and computational reliability, our approach can transform LLMs into more effective and reliable educational tools.
Code, models, and datasets can be accessed \href{https://github.com/luffycodes/Tutorbot-Spock-Phys}{here}
\footnote{\url{https://github.com/luffycodes/Tutorbot-Spock-Phys}}.

\section{Related Work}
In this section, we explore two critical areas that form the foundation of our study. 
First, we discuss the principles of ITS and how they seamlessly integrate with the concepts of Learning Science in the context of our stateful prompt design. 
Following this, we delve into the realm of LLMs, focusing on their mathematical abilities.

\subsection{Intelligent Tutoring Systems and Learning Science Principles}

ITS have carved a significant niche in the sphere of personalized education by providing students an interactive and individualized learning experience \cite{winkler2018unleashing}. 
ITS can be broadly classified into four categories \cite{feng2021systematic}. 
Firstly, Dialogue-Based ITS such as AutoTutor \cite{graesser2004autotutor} leverage natural language processing to pinpoint and rectify student misconceptions. 
Secondly, constraint-based modeling systems like KERMIT \cite{suraweera2002kermit} utilize predefined constraints to steer student learning. 
Thirdly, Knowledge Tracing models \cite{liu2022open, sonkar2020qdkt} track student knowledge states to capture their problem-solving skills. 
Lastly, Bayesian modeling \cite{sparfa} extends the model tracing approach by introducing Bayesian networks. 
Our proposed framework synergizes the principles of the first two types of ITS, utilizing a scaffolding strategy to deconstruct complex physics problems into smaller, manageable steps, and guiding students through these steps using conversational dialogues.

This scaffolding approach is deeply ingrained in specific learning science principles \cite{wing2006computational, shute2017demystifying}, which emphasize the efficacy of problem decomposition in fostering student learning. 
Our methodology aligns with the socio-constructivist model of learning \cite{vygotsky1978mind}, a model that champions scaffolding in education. 
This model advocates the breakdown of complex concepts into smaller subtasks, which are easier for learners to grasp—a strategy that is at the heart of our conversation design. 
Further, research indicates that optimal learning outcomes are achieved when the complexity of the task is synchronized with the learner's current abilities \cite{stone1998metaphor}. 
Thus, our approach merges the principles of ITS and learning science to provide an effective and engaging learning experience.

\subsection{Large Language Models and their Math abilities}

Recent advancements in NLP have led to the development of LLMs that show remarkable capabilities in generating human-like text and understanding complex language patterns. 
These capabilities make LLMs ideally suited for applications in ITS, which aim to engage students in a natural, interactive learning experience. 
Large-scale models such as GPT-4 \cite{gpt4sparkai} and PaLM \cite{chowdhery2022palm}, have garnered significant attention for their advanced capabilities. 
However, smaller models like LLaMA \cite{touvron2023llama,llama2} have also demonstrated promising results. 
These smaller models offer additional advantages such as increased customizability, safer deployment, and reduced costs.

Despite these advancements, one of the key challenges these models face is their limited accuracy in handling mathematical calculations. 
For instance, models like ChatGPT and GPT-4 have shown only $55\%$ and $59\%$ accuracy, respectively, on elementary tasks like three-digit multiplication \cite{gptbadmath}.
This limitation is a significant concern for their application in ITS, particularly for subjects like physics that often involve complex calculations. 
Various strategies have been explored to improve the mathematical capabilities of LLMs. Some of these approaches include the evol-instruct framework of WizardMath \cite{wizardmath}, combining LLMs with symbolic solvers \cite{symbolicmath}, or integrating them with external calculators \cite{calcmath}. 
Another innovative approach, which we adopt in our work, involves leveraging code \cite{pythonmath,pythonmath2} for solving simple math word problems \cite{gsm8k}.
Our unique contribution in this area is the introduction of `code soliloquies' that enables precise invocation of Python computations whenever a student's response necessitates it, thus significantly enhancing the computational reliability and interaction quality of the tutoring model.

\section{Methodology: Generating Conversations with Code Soliloquies}

In this section, we outline our innovative stateful prompt design, a methodology specifically developed to ensure that generated synthetic student-tutorbot conversations incorporate accurate calculations.
This design introduces `code soliloquies' - a novel feature that ensures the precise execution of computations during dialogues.

Our methodology employs two primary role prompts for the GPT-4 model, enabling it to simulate both the student and the tutorbot roles in generating these conversations.
While a straightforward approach might be to instruct the model to ‘act as a student/tutorbot and respond to the tutorbot/student’, we instead adopt a more sophisticated strategy inspired by the CLASS framework.

The student-specific prompt, detailed in appendix~\ref{app:student}, instructs GPT-student to generate inquiries and responses mimicking a real student's behavior.
The tutorbot-specific prompt instructs GPT-tutorbot to simplify complex problems into subproblems, helping the student solve these incrementally. 
The prompt also directs the bot to offer regular hints and not reveal answers prematurely.

The heart of our methodology lies in the intricate design of the tutorbot prompt because it is through this prompt design that we introduce the concept of `code soliloquies' - a critical feature ensuring the accurate execution of computations during dialogues.
The tutorbot prompt is composed of four sub-prompts representing four distinct states the tutorbot can be in.
These states form the backbone of the `code soliloquy'. 
They represent the internal thought process of the tutorbot as it determines whether a calculation is necessary for its response.

Now, let us delve into the specifics of each of these states to understand the tutorbot's internal monologue better.
\vspace{-2mm}
\begin{enumerate}
\setlength\itemsep{0em}
\item \textbf{Deciding State}: The `Deciding State' is the initial state in which the GPT-tutorbot determines whether a calculation is needed for its response to the student. In this state, the tutorbot is prompted with a specific prompt designed for this purpose (refer to appendix ~\ref{app:deciding_state}). The prompt instructs the model to make a binary decision - `yes' or `no' - to the question 'Use Python?', signaling whether Python computations are necessary.

The model should output `yes' if the student's response contains a numerical answer that needs verification using Python, or if the model anticipates its upcoming response to be reliant on mathematical calculations. Conversely, the model should output `no' when the scenario doesn't demand calculations. 

If the output is `yes', the tutorbot transitions to the `Use Python State'. If the output is `no', the conversation flow moves to the `No Python State'. Also, throughout this process, the Python functionalities remain hidden from the student.

\item \textbf{Use Python State}: If GPT-tutorbot model decides to use Python, the model is then prompted using prompt specific to `Use Python State' (refer to appendix ~\ref{app:use_python_state}).
The prompt instructs the model to first output a natural language description of the desired calculation, and then generate the corresponding Python code, enclosed within backticks and the `Python' keyword for easy parsing.

\item \textbf{Received Python State}: Post the execution of the Python code from previous step, the final state of code soliloquy is reached.
The GPT-tutorbot model is prompted using the specific prompt for this state (refer to appendix ~\ref{app:received_python_state}).
The model is instructed to use the Python output to assess the student's answer and provide suitable feedback.
If the student's answer is approximately close to the Python output (using rounding for comparison), the model is instructed to approve the answer. 
Once this state is concluded, the GPT-tutorbot's response is relayed to the GPT-student model, and the GPT-tutorbot model resets to the `Deciding State'.

\item \textbf{No Python State}: If the GPT-tutorbot concludes during the `Deciding State' that there's no need for Python, it transitions to the `No Python State'.
The prompt specific to this state (refer to appendix ~\ref{app:no_python_state}) instructs the model to continue the conversation and respond to the student.
Once the GPT-tutorbot's response is relayed to the GPT-student model, the GPT-tutorbot model reverts to the `Deciding State'. 
\end{enumerate}

Thus, our unique stateful prompt design facilitates the creation of synthetic conversations where the tutorbot, though limited in mathematical calculations ability, can skillfully use Python for computations to guide the conversation accurately.
This methodology significantly enhances the quality of synthetic dialogues, paving the way for the next crucial phase: training our model with these enriched dialogues.
In the following section, we discuss how these synthetic conversations are used to fine-tune our Higgs model.

\section{Dataset Curation}

Our methodology is underpinned by the careful curation of a high-quality dataset, termed PHY300, which comprises a diverse range of problems extracted from NCERT physics textbooks, covering topics from Newton's Laws to Thermodynamics to Electromagnetism.
A physics Subject Matter Expert (SME) was enlisted to provide solutions for these problems, incorporating the necessary mathematical computations within each solution.
This process yielded a diverse dataset of 300 unique question-solution pairs.

These questions-solution pairs become the seed for generating 450 mock student-tutorbot conversations, an instance of which is depicted in figure~\ref{fig:mock_con}.
For generating high quality conversations with rich pedagogical value, we further enrich SME provided solutions with the GPT-4 model.
The aim was to transform these succinct solutions into more comprehensive, step-by-step guides that explained the problem-solving process in a detailed, intuitive manner.
To achieve this, we designed a prompt that guided the GPT-4 model to not only elaborate on each step but also to articulate the underlying logic and principles guiding these steps, essentially providing a `teaching narrative'.
The output from GPT-4 was a comprehensive, easy-to-follow, and pedagogically sound step-by-step guide, based on the original SME-provided solution.
The exact wording of the prompt is provided in the appendix~\ref{app:solution_gen} for reference.

This process of enhancing the solutions leads us to a crucial assumption underpinning our methodology.
It's well-established that an LLM, on its own, struggles to solve physics problems independently.
This limitation is reflected in the LLM's performance on datasets like the MMLU \citep{mmlu} and GSM8K \citep{gsm8k}.
However, by providing the LLM with detailed, step-by-step solutions, we essentially equip it with a `script' that it can then translate into interactive, pedagogical dialogue.
This assumption is not unfounded but is based on the inherent capabilities of an LLM.
While the LLM may not independently generate solutions, it excels in natural language understanding and generation, making it well-suited to explain pre-determined solutions in an engaging and informative manner. 

Moreover, by leveraging Python for mathematical computations, we address the LLM's limitations in handling calculations.
During inference, we continue to hold this assumption, particularly for complex problems. 
For simpler problems, an LLM's ability to use Python computations might suffice without needing detailed solutions.

Thus our conversational dataset construction methodology enables the LLM to teach effectively by scaffolding the learning process by making informative use of GPT-enhanced dataset.
This sets the stage for the next critical phase: fine-tuning our Higgs model on this enriched dataset to develop a robust and effective tutoring system.

\section{Model Training}
In this section, we delve into the specifics of training our Higgs model, a fine-tuned version of the Llama-2-70b-chat base model \citep{llama2}.
We used LoRA \cite{lora} to freeze the base model weights, but only train rank decomposition matrices for each layer of the Llama model.
In addition to applying LoRA to the query, key, and value matrices, we also fine-tuned the projection and MLP matrices, a strategy known to boost performance \citep{qlora}.
The Higgs model was trained with an initial learning rate of $1e-4$ for $25$ epochs. 
We used a cosine learning rate scheduler and batch size of $16$.

The cost of training \botname can be broken down into two primary components.
First, the creation of conversational dataset involves prompting GPT-4, which costs approximately \$300 each. 
Second, we fine-tune the model using the CLM loss on conversational dataset for $25$ epochs.
This process is executed on $8$ NVIDIA RTX 48-GB A6000 GPUs and runs for three days.
In summary, the implementation of Higgs model involves conversational dataset generation using GPT-4, model selection, and domain-specific fine-tuning.

\section{Model Evaluation}
\label{sec:eval}
To accurately gauge the capabilities of our Higgs model, we designed an extensive evaluation protocol.
This protocol, equipped with crucial metrics, measures Higgs' ability to effectively utilize Python computations within an educational dialogue.
Paired with a Subject Matter Expert (SME) to provide measurements for each metric on a set of test questions, this protocol forms the backbone of our evaluation process.

\subsection{Evaluation Protocol}
Our evaluation protocol is designed to assess Higgs' performance in a comprehensive manner, focusing on its ability to appropriately use Python for calculations and its overall reliability in an educational dialogue.
The protocol is centered around four key performance metrics:

\vspace{-2mm}
\begin{enumerate}
\setlength\itemsep{0em}

\item Python Usage Accuracy: This first metric evaluates Higgs' ability to accurately determine when Python computations are needed within the dialogue. 
Specifically, it assesses whether Higgs correctly invokes Python to generate a suitable response or provide feedback to the student, for example in instances where the student's response includes a numerical answer requiring confirmation. 
This metric, therefore, serves as an indicator of Higgs' precision in recognizing and responding to calculation-dependent scenarios within the educational dialogue.

\item Non-Usage of Python: The second metric evaluates Higgs' ability to correctly identify instances where the use of Python is unnecessary. 
This ensures that the model judiciously invokes Python and can effectively distinguish between calculation-dependent and independent scenarios.
Thus, this metric complements the first one by evaluating Higgs' ability to correctly avoid the use of Python when it is not needed.

\item Code Compilation: The third metric gauges the reliability of the Python code generated by Higgs. This involves checking if the code is syntactically correct and if it compiles without errors. A successful compilation validates the model's capability to generate executable Python code.

\item Calculation Verification: The final and most critical metric measures Higgs' ability to verify calculations using Python. This assesses the model's competence in cross-verifying a student's calculation-based answer and providing accurate feedback.
\end{enumerate}

Each of these metrics is binary, indicating either a success (1) or a failure (0) for the given task. This comprehensive evaluation protocol allows us to thoroughly assess Higgs' performance, ensuring that it accurately and reliably utilizes Python computations in the context of an educational dialogue.

\vspace{-2mm}
\subsection{Preliminary SME Evaluation}
\vspace{-2mm}
\begingroup
\renewcommand{\arraystretch}{1.4} 
\begin{table*}[t!]
\centering
\normalsize
\resizebox{1.0\textwidth}{!}{
\begin{tabular}{c|c|c|c p{3cm}}
\hline
\textbf{Python Usage Accuracy} & \textbf{Non-Usage of Python} & \textbf{Code Compilation} & \textbf{Calculation Verification} \\ \hline
1.0 & 1.0 & 0.97 & 0.88 \\ \hline
\end{tabular}
}
\caption{
Performance metrics for Higgs model on a dataset of 50 test cases.
Each case is a physics question posed with both correct and incorrect answers. 
The model demonstrated flawless aptitude in identifying when to employ or bypass Python computations and generated syntactically correct Python code with near-perfect consistency. 
Despite a marginally lower score in verifying calculations, primarily due to challenges with equation rearrangement, the overall performance strongly asserts Higgs model's proficient and accurate use of Python in educational dialogues involving calculations.
}
\label{tab:eval}
\vspace{-4mm}

\end{table*}
\endgroup

Our evaluation protocol was executed in collaboration with a Subject Matter Expert (SME), who devoted six hours to this task. 
The SME tested our model on a set of 25 questions, covering a wide range of topics. 
Each question was introduced to the model twice, once with the correct answer and once with an incorrect answer.

The purpose of this methodology was two-fold. 
Firstly, when the correct answer was provided, we assessed if Higgs could accurately fact-check the answer using Python computations. 
Secondly, when the incorrect answer was provided, we evaluated whether Higgs could identify the error and provide the correct feedback to the student.

The results of this evaluation process are summarized in the evaluation table ~\ref{tab:eval}. 
The Higgs model demonstrated an impressive performance across all metrics, showcasing its ability to accurately and reliably utilize Python computations in an educational dialogue. 
The perfect score on `Python Usage Accuracy' and `Non-Usage of Python' demonstrates the model's exceptional ability to discern when Python computations are necessary or superfluous during the conversation.
The high score in `Code Compilation' indicates that the Python code generated by the model is almost always syntactically correct and executable.

The `Calculation Verification' score, while slightly lower than the others, is still notably high. 
This metric shows the model's ability to correctly verify student's answers using Python computations.
Our SME observed that the model struggled with questions that required equation rearrangement, a known limitation of the model's mathematical capabilities.
This observation provides context to the slightly lower score in this metric.

Overall, these scores affirm the effectiveness of our methodology and the resulting proficiency of the Higgs model.
Higgs' successful usage of Python computations significantly enhance the quality and accuracy of its educational dialogues, making it a powerful tool for AI-assisted education.

\section{Conclusion}
Our research presents a novel stateful prompt design that significantly bolsters the quality of synthetic conversation datasets, particularly for calculation-intensive subjects.
Using an inner monologue or code soliloquy in a GPT-4 simulated tutorbot, we enable it to decide when a response requires calculations, script the necessary Python code, and leverage the output to generated accurate responses and feedback as a tutorbot.
This innovative use of code soliloquy effectively mitigates GPT-4's known limitation in handling calculations, thereby improving its utility in 
generating mathematically accurate conversations.
Our model, named Higgs, fine-tuned on these mock conversations, demonstrates the effectiveness of our approach in training large language models to accurately perform computations within an educational dialogue.
Demonstrating an impressive ability to accurately and consistently deploy Python for computations, Higgs underscores the significance of integrating code soliloquies in the creation of synthetic dialogue datasets.
Thus, our research underscores the importance of incorporating code soliloquies in the generation of synthetic conversation datasets, paving the way for more accurate and computationally reliable Intelligent Tutoring Systems.

\section*{Acknowledgements}
This work was supported by NSF grants 1842378, ONR grant N0014-20-1-2534, AFOSR grant FA9550-22-1-0060, and a Vannevar Bush Faculty Fellowship, ONR grant N00014-18-1-2047.

\bibliography{custom}

\begin{thebibliography}{28}
\expandafter\ifx\csname natexlab\endcsname\relax\def\natexlab#1{#1}\fi

\bibitem[{Bubeck et~al.(2023)Bubeck, Chandrasekaran, Eldan, Gehrke, Horvitz, Kamar, Lee, Lee, Li, Lundberg et~al.}]{gpt4sparkai}
S{\'e}bastien Bubeck, Varun Chandrasekaran, Ronen Eldan, Johannes Gehrke, Eric Horvitz, Ece Kamar, Peter Lee, Yin~Tat Lee, Yuanzhi Li, Scott Lundberg, et~al. 2023.
\newblock {Sparks of Artificial General Intelligence: Early experiments with GPT-4}.
\newblock \emph{arXiv preprint arXiv:2303.12712}.

\bibitem[{Chen et~al.(2022)Chen, Ma, Wang, and Cohen}]{pythonmath2}
Wenhu Chen, Xueguang Ma, Xinyi Wang, and William~W Cohen. 2022.
\newblock Program of thoughts prompting: Disentangling computation from reasoning for numerical reasoning tasks.
\newblock \emph{arXiv preprint arXiv:2211.12588}.

\bibitem[{Chiang et~al.(2023)Chiang, Li, Lin, Sheng, Wu, Zhang, Zheng, Zhuang, Zhuang, Gonzalez, Stoica, and Xing}]{vicuna2023}
Wei-Lin Chiang, Zhuohan Li, Zi~Lin, Ying Sheng, Zhanghao Wu, Hao Zhang, Lianmin Zheng, Siyuan Zhuang, Yonghao Zhuang, Joseph~E. Gonzalez, Ion Stoica, and Eric~P. Xing. 2023.
\newblock \href {https://lmsys.org/blog/2023-03-30-vicuna/} {Vicuna: An open-source chatbot impressing gpt-4 with 90\%* chatgpt quality}.

\bibitem[{Chowdhery et~al.(2022)Chowdhery, Narang, Devlin, Bosma, Mishra, Roberts, Barham, Chung, Sutton, Gehrmann et~al.}]{chowdhery2022palm}
Aakanksha Chowdhery, Sharan Narang, Jacob Devlin, Maarten Bosma, Gaurav Mishra, Adam Roberts, Paul Barham, Hyung~Won Chung, Charles Sutton, Sebastian Gehrmann, et~al. 2022.
\newblock Palm: Scaling language modeling with pathways.
\newblock \emph{arXiv preprint arXiv:2204.02311}.

\bibitem[{Cobbe et~al.(2021)Cobbe, Kosaraju, Bavarian, Chen, Jun, Kaiser, Plappert, Tworek, Hilton, Nakano et~al.}]{gsm8k}
Karl Cobbe, Vineet Kosaraju, Mohammad Bavarian, Mark Chen, Heewoo Jun, Lukasz Kaiser, Matthias Plappert, Jerry Tworek, Jacob Hilton, Reiichiro Nakano, et~al. 2021.
\newblock Training verifiers to solve math word problems.
\newblock \emph{arXiv preprint arXiv:2110.14168}.

\bibitem[{Dettmers et~al.(2023)Dettmers, Pagnoni, Holtzman, and Zettlemoyer}]{qlora}
Tim Dettmers, Artidoro Pagnoni, Ari Holtzman, and Luke Zettlemoyer. 2023.
\newblock Qlora: Efficient finetuning of quantized llms.
\newblock \emph{arXiv preprint arXiv:2305.14314}.

\bibitem[{Dziri et~al.(2023)Dziri, Lu, Sclar, Li, Jian, Lin, West, Bhagavatula, Bras, Hwang et~al.}]{gptbadmath}
Nouha Dziri, Ximing Lu, Melanie Sclar, Xiang~Lorraine Li, Liwei Jian, Bill~Yuchen Lin, Peter West, Chandra Bhagavatula, Ronan~Le Bras, Jena~D Hwang, et~al. 2023.
\newblock Faith and fate: Limits of transformers on compositionality.
\newblock \emph{arXiv preprint arXiv:2305.18654}.

\bibitem[{Feng et~al.(2021)Feng, Magana, and Kao}]{feng2021systematic}
Shi Feng, Alejandra~J Magana, and Dominic Kao. 2021.
\newblock A systematic review of literature on the effectiveness of intelligent tutoring systems in stem.
\newblock In \emph{2021 IEEE Frontiers in Education Conference (FIE)}, pages 1--9. IEEE.

\bibitem[{Gao et~al.(2023)Gao, Madaan, Zhou, Alon, Liu, Yang, Callan, and Neubig}]{pythonmath}
Luyu Gao, Aman Madaan, Shuyan Zhou, Uri Alon, Pengfei Liu, Yiming Yang, Jamie Callan, and Graham Neubig. 2023.
\newblock Pal: Program-aided language models.
\newblock In \emph{International Conference on Machine Learning}, pages 10764--10799. PMLR.

\bibitem[{Graesser et~al.(2004)Graesser, Lu, Jackson, Mitchell, Ventura, Olney, and Louwerse}]{graesser2004autotutor}
Arthur~C Graesser, Shulan Lu, George~Tanner Jackson, Heather~Hite Mitchell, Mathew Ventura, Andrew Olney, and Max~M Louwerse. 2004.
\newblock Autotutor: A tutor with dialogue in natural language.
\newblock \emph{Behavior Research Methods, Instruments, \& Computers}, 36:180--192.

\bibitem[{He-Yueya et~al.(2023)He-Yueya, Poesia, Wang, and Goodman}]{symbolicmath}
Joy He-Yueya, Gabriel Poesia, Rose~E Wang, and Noah~D Goodman. 2023.
\newblock Solving math word problems by combining language models with symbolic solvers.
\newblock \emph{arXiv preprint arXiv:2304.09102}.

\bibitem[{Hendrycks et~al.(2020)Hendrycks, Burns, Basart, Zou, Mazeika, Song, and Steinhardt}]{mmlu}
Dan Hendrycks, Collin Burns, Steven Basart, Andy Zou, Mantas Mazeika, Dawn Song, and Jacob Steinhardt. 2020.
\newblock \href {http://arxiv.org/abs/2009.03300} {Measuring massive multitask language understanding}.
\newblock \emph{CoRR}, abs/2009.03300.

\bibitem[{Hu et~al.(2021)Hu, Shen, Wallis, Allen-Zhu, Li, Wang, Wang, and Chen}]{lora}
Edward~J Hu, Yelong Shen, Phillip Wallis, Zeyuan Allen-Zhu, Yuanzhi Li, Shean Wang, Lu~Wang, and Weizhu Chen. 2021.
\newblock Lora: Low-rank adaptation of large language models.
\newblock \emph{arXiv preprint arXiv:2106.09685}.

\bibitem[{Kasneci et~al.(2023)Kasneci, Se{\ss}ler, K{\"u}chemann, Bannert, Dementieva, Fischer, Gasser, Groh, G{\"u}nnemann, H{\"u}llermeier et~al.}]{kasneci2023chatgpt}
Enkelejda Kasneci, Kathrin Se{\ss}ler, Stefan K{\"u}chemann, Maria Bannert, Daryna Dementieva, Frank Fischer, Urs Gasser, Georg Groh, Stephan G{\"u}nnemann, Eyke H{\"u}llermeier, et~al. 2023.
\newblock Chatgpt for good? on opportunities and challenges of large language models for education.
\newblock \emph{Learning and Individual Differences}, 103:102274.

\bibitem[{Liu et~al.(2022)Liu, Wang, Baraniuk, and Lan}]{liu2022open}
Naiming Liu, Zichao Wang, Richard Baraniuk, and Andrew Lan. 2022.
\newblock Open-ended knowledge tracing for computer science education.
\newblock In \emph{Proceedings of the 2022 Conference on Empirical Methods in Natural Language Processing}, pages 3849--3862.

\bibitem[{Luo et~al.(2023)Luo, Sun, Xu, Zhao, Lou, Tao, Geng, Lin, Chen, and Zhang}]{wizardmath}
Haipeng Luo, Qingfeng Sun, Can Xu, Pu~Zhao, Jianguang Lou, Chongyang Tao, Xiubo Geng, Qingwei Lin, Shifeng Chen, and Dongmei Zhang. 2023.
\newblock Wizardmath: Empowering mathematical reasoning for large language models via reinforced evol-instruct.
\newblock \emph{arXiv preprint arXiv:2308.09583}.

\bibitem[{Shute et~al.(2017)Shute, Sun, and Asbell-Clarke}]{shute2017demystifying}
Valerie~J Shute, Chen Sun, and Jodi Asbell-Clarke. 2017.
\newblock Demystifying computational thinking.
\newblock \emph{Educational research review}, 22:142--158.

\bibitem[{Sonkar et~al.(2023)Sonkar, Liu, Mallick, and Baraniuk}]{sonkar2023class}
Shashank Sonkar, Lucy Liu, Debshila~Basu Mallick, and Richard~G Baraniuk. 2023.
\newblock Class meet spock: An education tutoring chatbot based on learning science principles.
\newblock \emph{arXiv preprint arXiv:2305.13272}.

\bibitem[{Sonkar et~al.(2020)Sonkar, Waters, Lan, Grimaldi, and Baraniuk}]{sonkar2020qdkt}
Shashank Sonkar, Andrew~E Waters, Andrew~S Lan, Phillip~J Grimaldi, and Richard~G Baraniuk. 2020.
\newblock qdkt: Question-centric deep knowledge tracing.
\newblock \emph{arXiv preprint arXiv:2005.12442}.

\bibitem[{Stone(1998)}]{stone1998metaphor}
C~Addison Stone. 1998.
\newblock The metaphor of scaffolding: Its utility for the field of learning disabilities.
\newblock \emph{Journal of learning disabilities}, 31(4):344--364.

\bibitem[{Suraweera and Mitrovic(2002)}]{suraweera2002kermit}
Pramuditha Suraweera and Antonija Mitrovic. 2002.
\newblock Kermit: A constraint-based tutor for database modeling.
\newblock In \emph{Intelligent Tutoring Systems: 6th International Conference, ITS 2002 Biarritz, France and San Sebastian, Spain, June 2--7, 2002 Proceedings 6}, pages 377--387. Springer.

\bibitem[{Touvron et~al.(2023{\natexlab{a}})Touvron, Lavril, Izacard, Martinet, Lachaux, Lacroix, Rozi{\`e}re, Goyal, Hambro, Azhar et~al.}]{touvron2023llama}
Hugo Touvron, Thibaut Lavril, Gautier Izacard, Xavier Martinet, Marie-Anne Lachaux, Timoth{\'e}e Lacroix, Baptiste Rozi{\`e}re, Naman Goyal, Eric Hambro, Faisal Azhar, et~al. 2023{\natexlab{a}}.
\newblock Llama: Open and efficient foundation language models.
\newblock \emph{arXiv preprint arXiv:2302.13971}.

\bibitem[{Touvron et~al.(2023{\natexlab{b}})Touvron, Martin, Stone, Albert, Almahairi, Babaei, Bashlykov, Batra, Bhargava, Bhosale et~al.}]{llama2}
Hugo Touvron, Louis Martin, Kevin Stone, Peter Albert, Amjad Almahairi, Yasmine Babaei, Nikolay Bashlykov, Soumya Batra, Prajjwal Bhargava, Shruti Bhosale, et~al. 2023{\natexlab{b}}.
\newblock Llama 2: Open foundation and fine-tuned chat models.
\newblock \emph{arXiv preprint arXiv:2307.09288}.

\bibitem[{Vygotsky and Cole(1978)}]{vygotsky1978mind}
Lev~Semenovich Vygotsky and Michael Cole. 1978.
\newblock \emph{Mind in society: Development of higher psychological processes}.
\newblock Harvard university press.

\bibitem[{Waters et~al.(2012)Waters, Lan, Studer, and Baraniuk}]{sparfa}
Andrew~E Waters, Andrew~S Lan, Christoph Studer, and Richard~G Baraniuk. 2012.
\newblock Learning analytics via sparse factor analysis.
\newblock In \emph{26th Annual Conference on Neural Information Processing Systems: Workshop Personalizing Education With Machine Learning (NIPS 2012)}.

\bibitem[{Wei et~al.(2022)Wei, Wang, Schuurmans, Bosma, Xia, Chi, Le, Zhou et~al.}]{calcmath}
Jason Wei, Xuezhi Wang, Dale Schuurmans, Maarten Bosma, Fei Xia, Ed~Chi, Quoc~V Le, Denny Zhou, et~al. 2022.
\newblock Chain-of-thought prompting elicits reasoning in large language models.
\newblock \emph{Advances in Neural Information Processing Systems}, 35:24824--24837.

\bibitem[{Wing(2006)}]{wing2006computational}
Jeannette~M Wing. 2006.
\newblock Computational thinking.
\newblock \emph{Communications of the ACM}, 49(3):33--35.

\bibitem[{Winkler and S{\"o}llner(2018)}]{winkler2018unleashing}
Rainer Winkler and Matthias S{\"o}llner. 2018.
\newblock Unleashing the potential of chatbots in education: A state-of-the-art analysis.
\newblock In \emph{Academy of management annual meeting (AOM)}.

\end{thebibliography}
\bibliographystyle{acl_natbib}

\appendix
\onecolumn
\section{Prompts}
\subsection{Prompt for GPT-4 to act as student}
\label{app:student}

\begin{lstlisting}[mathescape=true,basicstyle=\ttfamily\smalltonormalsize]
You are a high school student who's asking a tutorbot for guidance on a physics question. The question is "{question}".
    
The current conversation history is as follows:
"{history}"

Give only one response as the student. Do not generate Tutorbot responses. 

Generate an incorrect response 10% of the time. Use the following strategies to generate an incorrect response:
- apply the wrong formulae
- incorrectly rearrange the formulae to isolate the unknown variable on one side
- perform unit conversion incorrectly
- error in calculations
\end{lstlisting}

\subsection{Prompt for GPT-4 to act as tutorbot (deciding state)}
\label{app:deciding_state}
\begin{lstlisting}[mathescape=true,basicstyle=\ttfamily\smalltonormalsize]
You are a Tutorbot, an AI-powered chatbot designed to help students with a question by guiding the student step-by-step. Tutorbot helps the student by breaking down the main problem into steps, and helps the student solve each step sequentially. By default, Tutorbot can only provide hints. If the student asks for the answer or the student has answered incorrectly 3 times, provide the answer and move on to the next step to avoid getting stuck on a step.
If the student provides a numerical answer, Tutorbot generates Python code and uses it to verify the student's answer. If a mathematical calculation is required, Tutorbot generates Python code and uses the Python output to guide its response as the tutorbot. Utilize Python code as often as possible, especially to verify the student's calculations. Only verify the calculations within the student's most recent response. This Python functionality is hidden from the student.

The student's question is the following:
"{question}"

The step-by-step solution is formatted as "Step 1) ... Step 2) ... Step 3) ... etc." The step-by-step solution for the question is the following:
"{solution}"
Guide the student through the provided steps in the solution to solve the problem. The solution is hidden from the student.

The current conversation history is as follows:
"{history}"

Function of "Use Python" is to decide whether to use Python or not. Choose one:
y) Yes, use python. If student provided a mathematical answer, verify it by using Python, or if tutorbot think his next response relies on math calculations, get the output from Python
n) No, do not use python

If you choose to use Python ("y"), output the following JSON object, replacing any instances of ".." and following each field with a comma except for the last one:
{{
"Use Python": "y",
"Description": ".."
}}

The function of the "Description" field is to provide a natural language description of the desired calculation. Include all numerical values for all of the inputs. Always include the value of the student's calculation. Assume the description will be read out of context. Be as detailed as possible so that the description can be understood without context.

If you choose not to use python ("n"), output the following JSON structure:
{{
"Use Python": "n"
}}

Again, utilize Python code as often as possible. If the student provides a mathematical calculation in their most recent response, always verify it.
\end{lstlisting}

\subsection{Prompt for GPT-4 to act as tutorbot (use python state)}
\label{app:use_python_state}
\begin{lstlisting}[mathescape=true,basicstyle=\ttfamily\smalltonormalsize]
You are an AI-powered code generation bot. Given a natural language description of the desired calculation, generate the corresponding Python code.

The description of the desired calculation is the following:
"{description}"

Generate executable Python code with surrounding backticks and "python" keyword. Include comments. Do not use input() or print() statements. When comparing the student result and actual result, import the "math" module and use math.isclose() with rel_tol=0.01 by default. Only declare variables for inputs that have a numerical value. If a numerical value is not given for the student's result in the description, do not declare a variable, do not use a fake value for the student's result, and do not use math.isclose(). State the variables that store the results in the "Result Variable" field, separated by commas. Output everything in the following JSON object, following each field with a comma except for the last one:
{{
"Python":
{{
"Python Code": "``` python ..```",
"Result Variable": "Variable that the final answer is stored in"
}}
}}
\end{lstlisting}

\subsection{Prompt for GPT-4 to act as tutorbot (received python state)}
\label{app:received_python_state}
\begin{lstlisting}[mathescape=true,basicstyle=\ttfamily\smalltonormalsize]
You are a Tutorbot, an AI-powered chatbot designed to help students with a question by guiding the student step-by-step. Tutorbot helps the student by breaking down the main problem into steps, and helps the student solve each step sequentially. By default, Tutorbot can only provide hints. If the student asks for the answer or the student has answered incorrectly 3 times, provide the answer and move on to the next step to avoid getting stuck on a step. However, if there is no next step, mark the problem as finished.

The student's question is the following:
"{question}"

The step-by-step solution is formatted as "Step 1) ... Step 2) ... Step 3) ... etc." The step-by-step solution for the question is the following:
"{solution}"
Guide the student through the provided steps in the solution to solve the problem. The solution is hidden from the student.

The current conversation history is as follows:
"{history}"

The description of the Tutorbot's Python code was the following:
"{description}" 

The output from Tutorbot's Python code is the following:
"{python_output}"
Use Tutorbot's Python output to evaluate the student's answer and provide feedback to the student. If the student's answer is approximately close to the Python output (i.e. use rounding), then approve the answer and move on to the next step.

Put all the output in the following JSON structure, replacing any instances of ".." and following each field with a comma except for the last one:
{{
"Thoughts of Tutorbot": "..",
"Evaluation of Student Response": "a/b/c/d/e/f/g",
"Action Based on Evaluation": "1/2/3/4/5/6/7/8/9/10/11/12",
"Step Number": "..",
"Step State": "p/q/r/t",
"Tutorbot Response": ".."
}}
Decide only one "Evaluation of Student Response" and "Action Based on Evaluation" at a time.

The function of "Thoughts of Tutorbot" is to decide the student's response evaluation and the step state. It is a natural language description of what Tutorbot has decided.

Function of "Evaluation of Student Response":
a) Evaluating Incorrect Response
b) Evaluating Correct Response
c) Evaluating Partially Correct Response
d) Evaluating Ambiguous or Unclear or Short Response
e) Redirecting Off-topic Response
f) Responding to Student Inquiries
g) N/A

Tutorbot's "Action Based on the Evaluation":
If "a" is the evaluation, then:
1) Promptly notify the student about the mistake, Provide constructive feedback to
pinpoint the errors, Offer helpful hints
2) Step in to provide a solution if the student is unable to answer even after multiple
attempts.

If "b" is the evaluation, then:
3) Confirm the correct answer. Check for completeness for the answer to the step.
If the solution is incomplete, notify the student to complete the solution.

If "c" is the evaluation, then:
4) Acknowledge the accurate parts, Promptly notify the student about the mistake, Provide
constructive feedback to pinpoint the errors, Offer helpful hints
5) Step in to provide a solution if the student is unable to answer even after multiple
attempts.

If "d" is the evaluation, then:
6) Actively seek clarification through relevant follow-up questions. Request the student
to provide more specific information.

If "e" is the evaluation, then:
7) Skillfully redirect the student's attention to the subject matter. Provide guidance on
how to approach the question appropriately.

If "f" is the evaluation, then:
8) If student asks for help, provide a hint for the current step.
9) If student asks for a solution, give student the solution, marked current step
finished, and move to the next step.
10) If student asks to move to previous step, marked current step finished,
and move to the previous step.
11) If none apply, prioritize addressing the inquiry. Offer relevant support and guidance
to meet the student's specific needs.

If "g" is the evaluation, then:
12) N/A

Function of "Step Number" is to specify what the current step is according to the provided solution.

Function of "Step State" is to guide through steps:
p) N/A
q) One of the steps is currently being solved
r) Step finished, moving to next step that is not finished
t) Step finished, no next step, problem finished
\end{lstlisting}

\subsection{Prompt for GPT-4 to act as tutorbot (no python state)}
\label{app:no_python_state}
\begin{lstlisting}[mathescape=true,basicstyle=\ttfamily\smalltonormalsize]
You are a Tutorbot, an AI-powered chatbot designed to help students with a question by guiding the student step-by-step. Tutorbot helps the student by breaking down the main problem into steps, and helps the student solve each step sequentially. Tutorbot can only provide hints. Only provide the answer when the student asks for it or the student has answered incorrectly 3 times. 

The student's question is the following:
"{question}"

The step-by-step solution is formatted as "Step 1) ... Step 2) ... Step 3) ... etc." The step-by-step solution for the question is the following:
"{solution}"
Guide the student through the provided steps in the solution to solve the problem. The solution is hidden from the student.

The current conversation history is as follows:
"{history}"

Put all the output in the following JSON structure, replacing any instances of ".." and following each field with a comma except for the last one:
{{
"Thoughts of Tutorbot": "..",
"Evaluation of Student Response": "a/b/c/d/e/f/g",
"Action Based on Evaluation": "1/2/3/4/5/6/7/8/9/10/11/12",
"Step Number": "..",
"Step State": "p/q/r/t",
"Tutorbot Response": ".."
}}
Decide only one "Evaluation of Student Response" and "Action Based on Evaluation" at a time.

The function of "Thoughts of Tutorbot" is to decide the student's response evaluation and the step state. It is a natural language description of what Tutorbot has decided.

Function of "Evaluation of Student Response":
a) Evaluating Incorrect Response
b) Evaluating Correct Response
c) Evaluating Partially Correct Response
d) Evaluating Ambiguous or Unclear or Short Response
e) Redirecting Off-topic Response
f) Responding to Student Inquiries
g) N/A

Tutorbot's "Action Based on the Evaluation":
If "a" is the evaluation, then:
1) Promptly notify the student about the mistake, Provide constructive feedback to
pinpoint the errors, Offer helpful hints
2) Step in to provide a solution if the student is unable to answer even after multiple
attempts.

If "b" is the evaluation, then:
3) Confirm the correct answer. Check for completeness for the answer to the step.
If the solution is incomplete, notify the student to complete the solution.

If "c" is the evaluation, then:
4) Acknowledge the accurate parts, Promptly notify the student about the mistake, Provide
constructive feedback to pinpoint the errors, Offer helpful hints
5) Step in to provide a solution if the student is unable to answer even after multiple
attempts.

If "d" is the evaluation, then:
6) Actively seek clarification through relevant follow-up questions. Request the student
to provide more specific information.

If "e" is the evaluation, then:
7) Skillfully redirect the student's attention to the subject matter. Provide guidance on
how to approach the question appropriately.

If "f" is the evaluation, then:
8) If student asks for help, provide a hint for the current step.
9) If student asks for a solution, give student the solution, marked current step
finished, and move to the next step.
10) If student asks to move to previous step, marked current step finished,
and move to the previous step.
11) If none apply, prioritize addressing the inquiry. Offer relevant support and guidance
to meet the student's specific needs.

If "g" is the evaluation, then:
12) N/A

Function of "Step Number" is to specify what the current step is according to the provided solution.

Function of "Step State" is to guide through steps:
p) N/A
q) One of the steps is currently being solved
r) Step finished, moving to next step that is not finished
t) Step finished, no next step, problem finished
\end{lstlisting}

\subsection{Comprehensive solution generation prompt}
\label{app:solution_gen}

\begin{lstlisting}[mathescape=true,basicstyle=\ttfamily\smalltonormalsize]
Given a textbook problem and its textbook solution, generate a more descriptive, confluent step-by-step solution, explaining each step with natural language. Number each step with "Step 1) ... Step 2) ... Step 3) ... etc." Provide the intuition behind each step, as if you are a teacher explaining the solution to a student.
Divide steps up to be small and digestible. Steps can be of the following types but are not limited to:
- stating the known values that are provided in the question
- describing which equation(s) to use
- rearranging the equation to isolate the unknown variable on one side
- performing unit conversions
- plugging the known values into the equation and solving for the unknown variable
- etc.

The textbook problem is the following:
"{question}"

The solution is the following:
"{solution}"

Put the output in the following JSON structure, replacing any instances of ".." and following each field with a comma except for the last one:
{{
"Detailed Solution": "Step 1) .. Step 2) .. Step 3) .. etc.",
"Solution Outline": "Step 1) .. Step 2) .. Step 3) .. etc."
}}

The function of the "Detailed Solution" field is to provide a detailed step-by-step solution based on the provided solution. Explain each step in detail, providing in-depth reasoning behind each step, and include both equations and calculations. However, do not perform your own calculations-only use the calculations provided in the solution. The detailed solution should be longer and more thorough than the overview.

The function of the "Solution Outline" field is to create a low-level and concise outline of the detailed solution. Briefly explain each step and how they are connected in natural English. Include any equations or formulas and explain their purpose, but do not include any calculations. Describe the relationship between equations to demonstrate how the student can move from one step to the next.
\end{lstlisting}

\end{document}